# SMDDH: Singleton Mention detection using Deep Learning in Hindi Text

Kusum Lata*

Computer Science & Engineering, National Institute of Technology, Hamirpur, (HP), India, kusumlata@nith.ac.in

Pardeep Singh

Computer Science & Engineering, National Institute of Technology, Hamirpur (HP), India, pardeep@nith.ac.in

Kamlesh Dutta

Computer Science & Engineering, National Institute of Technology, Hamirpur (HP), India, kd@nith.ac.in

Mention detection is an important component of coreference resolution system, where mentions such as name, nominal, and pronominals are identified. These mentions can be purely coreferential mentions or singleton mentions (non-coreferential mentions). Coreferential mentions are those mentions in a text that refer to the same entities in a real world. Whereas, singleton mentions are mentioned only once in the text and do not participate in the coreference as they are not mentioned again in the following text. Filtering of these singleton mentions can substantially improve the performance of a coreference resolution process. This paper proposes a singleton mention detection module based on a fully connected network and a Convolutional neural network for Hindi text. This model utilizes a few hand-crafted features and context information, and word embedding for words. The coreference annotated Hindi dataset comprising of 3.6K sentences, and 78K tokens are used for the task. In terms of Precision, Recall, and F-measure, the experimental findings obtained are excellent.

.

## 1. INTRODUCTION

Coreference Resolution (CR) is the process to recognize the set of mentions, those point to the same entities in the text. The Coreference Resolution task consists of two main steps: 1) Mention detection module detects all mentions (entities) present in the text. 2) Clustering of mentions pointing to the same real-world entity in the text. Coreference resolution is an essential task that facilitates the variety of downstream applications of natural language processing such as Machine Translation [1,2], Text Summarization[3,4], Reading Comprehension [5,6], Entity linking [7,8], Question Answering [9,10], Sentiment Analysis [11,12]. The term "mention" has been commonly used by the researchers to represent entities specifying the occurrences listed in the document, which can be a name, nominal or a pronominal Florian et al. [13]. We explain the concept of mention with the help of following example:

SH1[1]: *फिल्म महोत्सव* में *प्रकाश झा* की *नई फिल्म अपहरण* का भी प्रीमियर होना है / *गंगाजल* के बाद *उसकी यह* किसी अलग विषय पर बनी *दूसरी फिल्म* है ।

SE1[2]: **Prakash Jha**'s **new film Apaharan** is also to premiere at **the film festival**. This is **his second film** on a different subject after **Gangajal**.

---

* ranapoo@gmail.com
[1] SH indicating sentence written in Hindi.
[2] SE indicating sentence written in English.

SHI1[3]: **Prakash Jha** kee **naee film apaharan** ka bhee **film mahotsav** mein preemiyar hona hai. **Gangaajal** ke baad **usakee** yah kisee alag vishay par banee **doosaree film** hai.

In this example of the sentence (in Hindi) , SH1,फिल्म महोत्सव (film festival /film Mahotsav), प्रकाश झा (Prakash Jha), नई फिल्म (naee film /new film), अपहरण (apaharan), उसकी (his /usakee), यह (this /yah), दूसरी फिल्म (second film /doosaree film), गंगाजल (Gangajal) are the mentions in this sentence. फिल्म *महोत्सव* (film Mahotsav), नई *फिल्म* (naee film), दूसरी *फिल्म* (doosaree film) are nominal mentions. प्रकाश झा (Prakash Jha), अपहरण (apaharan), गंगाजल (Gangaajal) are named mentions. *उसकी* (usakee), *यह* (yah) are pronominal mentions.

The coreference resolution process attempts to identify the cluster of mentions referring to the same entity. The coreference resolution process, for example, SH1, is shown in Fig.1.

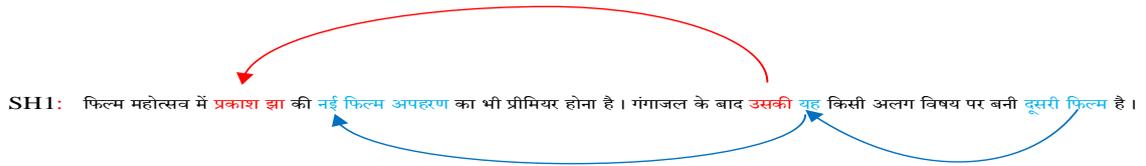

Fig.1. Coreference resolution process.

Further, these mentions were divided into two categories by Recasens et al. [14] as purely coreferential mentions and singleton mentions (non-coreferential mentions) as shown in Fig.2. The first coreferent mentions are those that are co-refered (point to the same entity in the real world). The second one, Singleton mentions exist only once in the text and does not participate in the coreference cluster. As shown in example SH1, {प्रकाश झा *(Prakash Jha)*, उसकी *(his /usakee)*} are in the cluster because both are pointing to the same entity. {नई फिल्म *(naee film /new film)*, अपहरण *(apaharan)*, यह *(this /yah)*, दूसरी फिल्म *(second film /doosaree film)*} are in the same cluster because they point to the same entity. फिल्म *महोत्सव* (film festival /film Mahotsav) and गंगाजल (Gangajal) are singleton mentions in the example SH1 as both are mentioned once and do not participate in the coreference process.

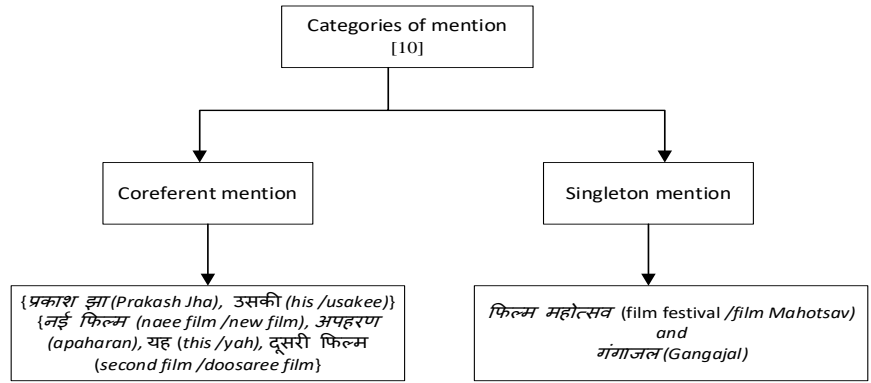

Fig.2. Category of mention according to frequency of reference.

---

[3] SHI indicating sentence written in Hinglish/ Roman Hindi (Hindi to English) for understanding.

Let us consider, 'M' be the set of mentions extracted from text using mention detection module  M = $\{M_1, M_2, M_3, \ldots, M_n\}$. Let, 'S' be the set of the mentions which are singleton. We assume that  singleton set, S = $\{S_1, S_2, S_3, \ldots, Sn\}$ where '$S_i$' is a  number of binary values indicating the presence of  mention Mi as a singleton by '1' and '0' for non-singleton mentions. For example, if  a document has 4 mentions, M= $\{M_1, M_2, M_3, M_4\}$  and singleton set S= $\{1,0,0,1\}$   then it implies that $M_1$ and $M_4$ are singleton mentions. Let, 'C' be the set of the mentions which are coreferent i-e, $C = \text{count}(M) - \text{count}(S)$. M set is the union of singleton and corefrent mention  i-e, $M = S \cup C$. The ratio of $M - S$ over $M$  is denoted by 'R' as shown in equation (1)

$$R = \frac{M-S}{M}$$ ………. (1)

If no singleton mention exist in the document i-e., $S = M - C = \emptyset$ , then R=1 implies that  all are coreferent mentions and theire is no need of performing singleton detection as all mentions are participating in coreference resolution. On the other hand , if substantial number of singltons exist in the document, they need to be filtered  before applying coreference resolution task as these singleton entities do not contribute in the coreference resolution but only increase the  search and space time. Therefore, in order to prune the search space while coreference resolution identification of singleton mentions is required.  We can say that the detection of singleton mention is the pre-processing step  for the CR task. The singleton detection module is used to detect the singleton mentions from the predicted mentions (mention produced by mention detection module) in the text. Kübler and  Zhekova [15] investigated the impact of singletons on the evaluation of CR systems. The filtering out singleton mention from predicted mentions reduces the search space of the CR system, which helps to improve the performance of the CR system.

A lot of work for languages like English, Russian, Indonesian, Arabic, etc has been done. On the other hand, Hindi language, being a Scarce-resource Language (SRL) due to the scarcity of the required computational linguistic resources such as good quality datasets, coreference resolution system, machine translation tools etc.,  inherently posses various challenges  such as variation in spelling, arrangement of words, restricted linguistic resources, when dealing with Hindi text as  explained by  the researchers [16–18]. Due to the scarcity of linguistic resources in Hindi, highly accurate methods such as Mention Detection, Singleton Mention Detection, and CR have been challenging to build. The creation of hand-crafted rules takes time for language specialists, and it necessitates a thorough understanding of language grammar. The machine learning-based methods also rely on in-depth knowledge of language grammar and hand-crafted features. Because of the lack of linguistic resources in Hindi, hand-crafting features are challenging. Deep learning-based models have recently demonstrated significantly improved outcomes in dealing with these issues. The authors [19–21]demonstrated the use of a deep learning-based model in the NLP task and discussed several deep learning models such as RNN [22], LSTM [23], CNN [24,25], Capsule Network [26,27], etc. The deep learning-based model has proven state-of-the-art performance in a variety of NLP applications, including such as Named Entity Recognition [28,29], Text classification [30], Text summarization [3,4], Chatbot [31,32], Machine Translation [1,2], Relation Extraction [33]. Deep neural network-based models have already been suggested to build the Singleton Mention Detection module in various languages, including English [34,35], Indonesian[36], etc. and they performed well.

The singleton mentions are approximately 50% in Hindi dataset used for our task; this implies that  $S \neq \emptyset$ and the value of 'R' as shown in equation (1) is not '1' and need to remove these singleton mentions from mention list available in the Hindi text. This work presents a singleton classifier using deep learning techniques to remove the singleton mentions from Hindi text. We evaluate the performance of the Singleton Mention Detection module for the Hindi language, which is helpful for the coreference resolution system, which we believe has  not  been tried so far A Convolutional Neural network (CNN) network model is utilized for the Singleton Mention Detection task for the presented work.

The contribution of our work in this paper is as follows:
- We present a neural network-based Singleton Mention Detection module to detect singleton mentions in Hindi text by utilizing CNN and word embedding.
- Our model aims to reduce the dependency on hand-crafted features.
- The proposed approach is evaluated on the Hindi Coreference dataset, and the performance of the model is evaluated

with different Hyper-parameters.

The remainder of the paper is divided into the following sections. Section 2 provides a thorough overview of the models for Singleton Mention Detection that have been developed, or the Related Work that has been done in the domain. The Proposed Approach for the work is described in Section 3. The Experimental Evaluation will be discussed in Section 4, and the Conclusion and Future Scope of our work will be discussed in Section 5.

## 2. RELATED WORK

This section presents the research work related to the identification of singleton mention in different languages. The singleton detection classifier task has received a lot of attention in the literature, prominently for English. Raghunathan et al. [37] employed a rule-based approach to eliminate erroneous mentions such as numeric entities. Björkelund and Farkas [38] employed binary MaxEnt classifiers to detect referential (participate in coreference cluster) and non-referential (do not involve in coreference cluster) for the pronouns *it, we, and you*.

Firstly, Recasens et al. [14] have used linguistic insights about how syntactic and semantic characteristics determine discourse entity lifespans to create a logistic regression model for predicting the singleton mentions. They evaluated the model on CoNLL-2012 Shared Task data. The accuracy of a model is 78 percent, but when combined with a Rule-based Standford coreference resolution system [39], it delivers a considerable improvement.

De Marneffe et al. [40] have extended the work described by Recasens et al. [14]. They utilized both surface features and vast numbers of linguistic features. They presented a considerably more in-depth analysis of various aspects of the lifespan model and examined how surface properties influence it. They evaluated the lifespan model's utility in predicting which phrases will serve as anchors in bridging anaphora. They also integrated the perfect lifespan model into the Berkeley coreference system [41]. The learning-based model utilized the log linear model and compared the results with prior results on the rule-based Stanford coreference system. They also did error analysis to understand better the errors that the life span model can help with.

Haagsma [34] developed a singleton detection system utilizing a neural network and word embeddings. A recursive autoencoder and a multi-layer perceptron are the two primary components of the singleton detection system. The recursive autoencoder is used to build fixed-length representations for multi-word mentions based on word embeddings. The actual singleton detection is done with a multi-layer perceptron. They evaluated the singleton model's performance on CoNLL-2012 Shared Task data, and accuracy is reported as 79.6%. They have also shown that while the performance of the Stanford coreference resolution system increases dramatically, the performance of the Berkeley coreference resolution system does not. The findings suggest that neural networks and word embeddings can improve both singleton detection and coreference resolution system.

Moosavi and Strube [42] utilized anchored SVM [43] with a polynomial kernel of degree two for classification. They chose a simple and small set of shallow features as compared to the set of features used by De Marneffe et al. [40]. The authors evaluated system with three configurations: S*urface* which used only shallow features, *Combined* used shallow features and linguistics features as utilized by De Marneffe et al. [40], but included only single features and *Confident* used shallow features along with high confidence prediction of SVM on CoNLL-2012 Shared Task data. The accuracy is reported. They showed that the performance of CR tasks could be enhanced by utilizing search space pruning. Surface, Combined, Confident accuracy is reported as 85.73%, 85.85%, 61.08%, respectively. They showed that by just utilizing shallow features, they obtained a new state-of-the-art performance for singleton mention detection that outperformed the results of De Marneffe et al. [40] for classifying both coreferent and non-coreferent mentions by a significant margin.

Wu and Ma [44] utilized *word, dependency, string, numeric, and mention* as input. They employed CNN to produce word representation by using word embeddings of each word in mention and FCN for getting representation for numeric, dependency, string features of mention. They evaluated the model on the CONLL-2012 shared task dataset and showed that the performance of CR task is increased by singleton classification, but separately, they did not evaluate singleton classifier.

Li et al. [35] developed a singleton mention detection by utilizing a multi-window and multi-filter convolutional neural network (MMCNN). The MMCNN model extracts multi-granularity level features and adequate information for coarser features to detect singleton mentions with fewer hand-designed features and more sentence information. They used logistic regression classifier to identify that whether the mention is a singleton or not. They evaluated the performance of the system on the English portion of the CoNLL 2012 Shared Task dataset. The Precision, Recall, F1-measure are reported as 81.15%, 87.27%, 84.10% respectively on dev data set and similarly, 78.35% 88.22%, 83.00% respectively on test dataset.

There are other languages in which work is done to identify singleton such as Russian, Indonesian, etc. Ionov and Toldova [45] developed a model based on a Random Forest classifier for detecting singleton mention in Russian texts. They examined several morphosyntactic and lexical characteristics, some of which had previously been employed for similar tasks in English, and proposed additional features based on discourse analysis. They found that the classifier's performance is lower than that of English. This classifier's Precision, Recall, F-measure with all features are reported as 60.0%, 70.80%, and 65.0%, respectively. They showed that the performance of coreference resolution classifiers could be improved by pruning singleton mention.

Auliarachman and Purwarianti [36] developed deep neural network models by utilizing a Convolutional neural network and Fully Connected Network (FCN) for identifying singleton mentions in Indonesian text. They used a set of features such as mention's word, context, mention features. The F-1 score for Non-Singleton, Singleton is reported as is 64.24%, 94.42%, respectively. They examined that if the singleton classifier performs well, the singleton exclusion procedure can significantly enhance the performance of coreference resolution systems. Even a low-performing trained singleton detection model increased a system's benefit from 29.55% to 22.76%.

Nowadays, NLP applications are also available for the Hindi language; as a result, a cutting-edge Hindi Singleton Mention Detection system is needed to bring out the new achievement and better results of these applications. Deep learning-based approaches do not appear to have been extensively investigated for Hindi text. Still, a few state-of-the-art deep learning-based Singleton Mention Detection models for other languages, such as English, Indonesian, and others, have been developed. The available system for Hindi Mention Detection used traditional approaches such as hand-crafted features. Only, Vasantlal [46] developed Rule-based-Mention Detection for Hindi text and utilized the hand-crafted features. Still, the author did not develop a system to detect singleton mention from the list of mentions generated by Rule-based-Mention Detection. Therefore, a Deep learning-based Singleton Mention Detection model for Hindi language text could be developed, which might provider cutting- edge results without the need for hand-crafted features.

## 3. PROPOSED APPROACH

This section describes our proposed approach used for detecting singleton mentions. After detecting all the mentions from Hindi text, removing singleton mentions can improve the performance of NLP applications such as CR task. We propose a deep learning-based singleton mention detection model which employ a convolutional neural network (CNN) or Long short-term memory (LSTM). As far as we know, we are the first to use a deep neural network to create a Singleton Mention Detection module for the Hindi language. Moosavi and Strube [42] utilized features such as *lemmas of mention's words, lemmas of context words (two next and previous words of mention), part of speech tags of each words in mention, part of speech tag of previous and next two words of mention, mention'words, mention type, binary features: whether mention appears again in the text, whether mention head appears in text again*. Auliarachman and Purwarianti [36] employed features such as *mentions' word embedding, context embedding, which include 10 previous words and 10 next words of mention, mention features* for singleton classifier using deep learning techniques.

The proposed Singleton Mention Detection model is a tweaked version of the model developed by Auliarachman and Purwarianti [36]. The proposed approach for singleton mention detection captures word-level features from raw text automatically. Fig. 3 depicts the block diagram of the proposed approach.

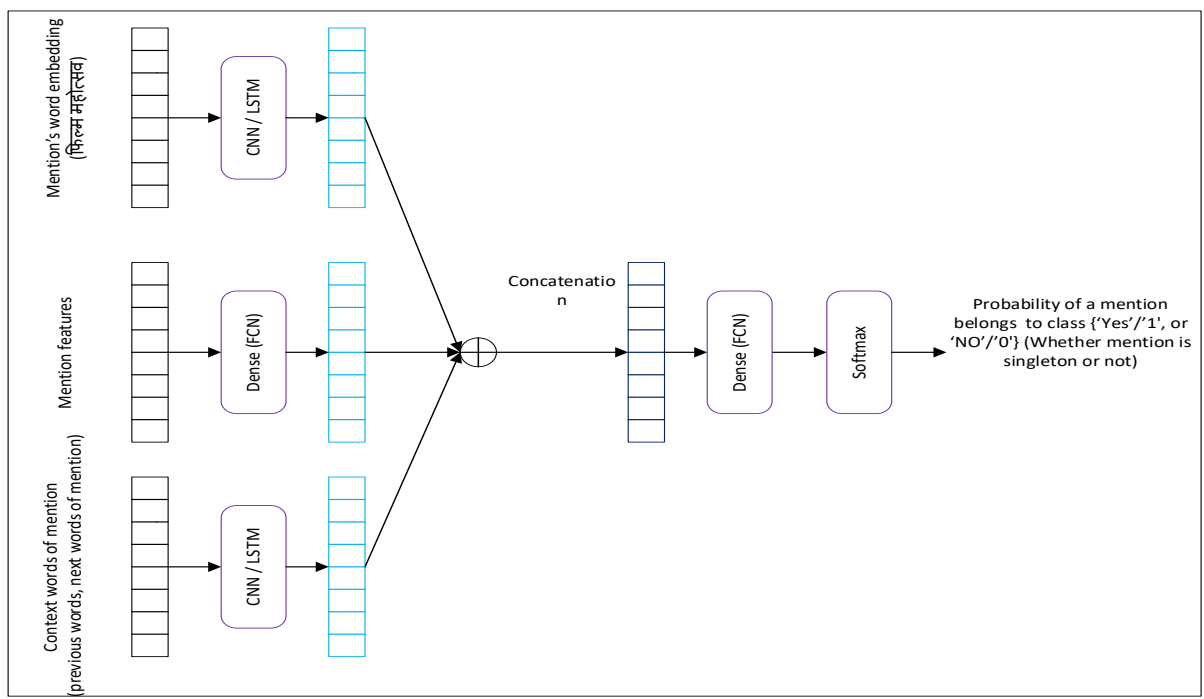

Fig. 3. Block diagram of proposed Singleton Mention Detection Module.

Moosavi and Strube [42] used lemmas and part of speech tags of each word in mention, lemmas of previous and next words of mentions, but our approach use word embeddings of mention's words, word embedding of context words obtained from pretrained word2vec. Auliarachman and Purwarianti [36] used 10 previous words and 10 next words of mention as context mention; however, our approach used two previous words and two next words of a mention as context.

The inputs for our proposed singleton detection module are described below:
- **Mention's word embedding:** we employ a pre-trained word embedding method to get word-level features in vector form.
- **Context embedding:** Two previous words of a mention and two next words after the mention are used as context. Pre-trained word embedding is used to represent context words in word vector form.
- **Mention' syntactic features**: We employ four programmatically produced binary features for this feature group, as shown in [1]. This group include (1) whether the mention is a pronoun or not, (2) whether the mention is a proper name or not, and (3) whether the mention is the first-person pronoun or not

The algorithm of the proposed model is described in the following, and the proposed model's various layers are detailed in the following subsections. Our approach consist of the following layer:
- Embedding Layer
- Convolution neural network (CNN)
- Fully Connected Network (FCN)
- Softmax

**Algorithm**: Detection of singleton mention from list of mentions

---

**Input**: List of mentions: Ms
**Output**: Singleton or Non singleton p(Ts), i-e Si={1 or 0}
**Dataset**: Training Set Ts
**Maximum length of mention M**: maxm_len
**Vector Dimension**: dims
**Maximum number of words in the word2vec model**: Maxm_NB_words
**Maximum number of words in the training set Ts**: $N_{word}$
**Step 1:** getwordvector (word2vec model)
    for each word $W_i$ in Ts do           where i=1,2,3.......$N_{word}$
        if word $w_i \in$ pre-trainedword2vec then
            word_vector$_{i*dims}$ ← pre-trainedword2vec ($W_i$)
        else
            word_vector$_{i*dims}$ ← Zero_vector initialization(*dims*)
        end if
    end for
**Step 2:** get_embedding_matrix (word2vec model)
    Index_Word = get Index values from pre-trainedword2vec
    for each word $W_i$, $k$ in Index_Word do
        if $k$>=Maxm_NB_words in word2vec model then
            continue
        try
            *embedding_vectors* ← getwordvector ($W_i$)
            *embedding_matrixx[k]* ← embedding_vectors
        except:
            pass
    end for
    return *embedding_matrixx*
**Step 3:** CNN(M):
    $M_W$ ← [$W_1, W_2, W_3, ........................., W{maxm_{len}}$]     Where $W_i \in M$, i=1,2, 3,....maxm_len
    $M_{word\_embedding\_layer\ (maxm\_len\ *\ dims)}$ ← Embedding ($M_W$, weights=embedding_matrixx)
    $M_{cnn\_layer1}$ ← Conv2D ($M_{word\_embedding\_layer\ maxm\_len*dims}$, *filter_size=2, filters=64, activation='relu'*)
    $M_{cnn\_layer2}$ ← Conv2D ($M_{maxm\_len*dims}$, *filter_size=3, filters=64, activation='relu'*)
    $M_{cnn\_layer3}$ ← Conv2D ($M_{maxm\_len*dims}$, *filter_size=4, filters=64, activation='relu'*)
    $M_{cnn\_maxpool1\ (max*dim1)}$ ← MaxPool2D ($M_{cnn\_layer1}$)
    $M_{cnn\_maxpool2\ (max*dim2)}$ ← MaxPool2D ($M_{cnn\_layer2}$)
    $M_{cnn\_maxpool3\ (max*dim3)}$ ← MaxPool2D ($M_{cnn\_layer3}$)
    $M_{concatenated}$ ← Concatenate ($M_{cnn\_maxpool1}, M_{cnn\_maxpool2}, M_{cnn\_maxpool3}$)
    $M_{w\_vec}$ ← Flatten ($M_{concatenated}$)
    $M_{w\_vec}$ ← Dense (*16, activation='relu'*)
**Step4:** for each in Ts do
    if (words_feature):
        $cnn_1$ = CNN (input_length = 10, vocab_size = len(embedding_matrixx), vector_size =
            len (embedding_matrixx [0]), embedding_matrixx = embedding_matrixx, filter_size=[2,3,4],
            num_filter = 64, trainable_embedding = False, output_size = 16)
    if (context_feature (Prev_word, Next _word)):
        $cnn_2$ = CNN (input_length = 10, vocab_size = len(embedding_matrixx), vector_size =
            len (embedding_matrixx [0]), embedding_matrixx = embedding_matrixx, filter_size=[2,3,4],
            num_filter = 64, trainable_embedding = False, output_size = 16)
    if (syntactic_feature):
        $Syn_W$ = Dense (*num_syntactic_features, layers = [32,16], dropout = 0.2*)
    $M_{word\_rep}$ = Concatenate ($cnn_1, Syn_W, cnn_2$)
    $M_{word\_rep}$ = Dense ($M_{word\_rep}$, *layers =[32, 8], dropout=0.2*)
    P(Ts) ← Dense ($M_{word\_rep}$, *units=2, activation=" softmax"*)
    end for

---

## 3.1 Embedding layer

### Word Embedding

The first step of our approach is to transform each word into vector form. The pre-trained word2vec is used to get the vector form of the word because Hindi is a resource-constrained language with little training data. The Wikipedia corpus is used to train Word2vec[4]. Word2vec employ a neural network model to learn wordvectors from a large corpus. We used pre-trained Word2vec, which is based on [47]. Therefore, word embedding uses to represent complete mention (each word in mention). The overfitting of the artificial neural network can be reduced effectively by using the pre-trained model. The vector size of each word is of dimension "$d_v$" and vocabulary size of pretrained word2vec is "v." The embedding matrix obtained is of $d_v * $ v size. The word vectors of each word in mention are combined together which is given to embedding layer which provide the word embedding of mention's words is a matrix of $l_w*d_v$ for each mention where $l_w$ is equal to the number of words in the mention and $d_v$ is the dimension of the vector.

### 3.2 Convolution Neural Network

**Convolution Layer and Max-pooling Layer**

The word embedding is fed as an input to the two-dimensional- CNN that scan various words and extracts information from words. This layer produces the word embedding of mention's words. This convolution layer employs filters such as filter size=2,3,4 for scanning tasks and generates a word vector corresponding to each filter. In addition, each vector is subjected to a max-pooling process to extract the essential characteristics, as shown in Fig. 4. After applying CNN, the output is the CNN representation of the mention's word. Similarly, the context feature of mention, which includes the two previous words and two next words of mention, is considered as input. The word embedding of context feature is obtained from pre-trained word2vec and given as input to the CNN layer, which produces the CNN representation for vector representation of the context of mention as shown in Fig. 4.

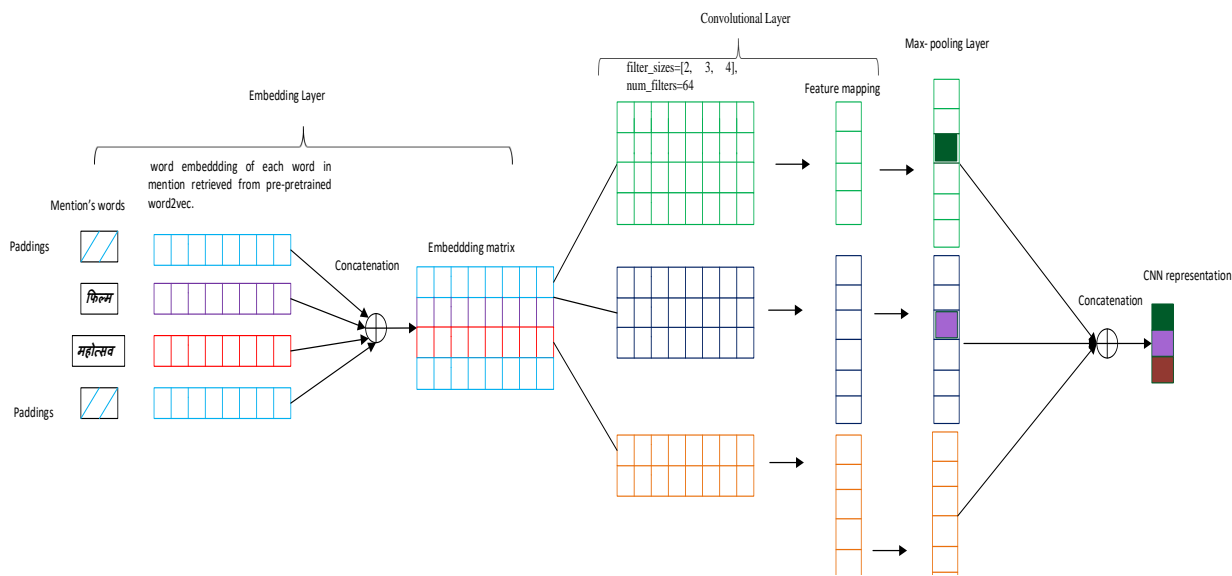

Fig. 4. CNN representation of mention's words.

### 3.3 Fully Connected Layer

The binary features of mentions are given to the Dense layer, also called a Fully Connected Network (FCN). Every neuron in this layer is connected to every other neuron in the previous layer. In traditional neural network models, the connection topology is similar to layers. The dropout regularization is performed to the fully connected layer for reducing overfitting and enhancing generalization performance. Neurons that are dropped have a chance of being temporarily removed from the network during training. Dropped neurons are disregarded when computing the input and output for both forward and backward propagation. As a result, the dropout approach inhibits neurons from over-co-adapting by making any neuron's existence unreliable. The function used by the Dense layer (FCN) for the activation is 'RELU' (*Rectified linear unit*) which is defined as :

$$f_n(z) = \begin{cases} z, & \text{if } z > 0 \\ 0, & \text{otherwise} \end{cases} \quad\quad\quad\quad\quad\quad (2)$$

and after this, three vector representations are concatenated to produce a single input and fed as an input to FCN.

---

[4] https://github.com/Kyubyong/wordvectors

### 3.4 Softmax Layer

The output produced by a fully connected layer is considered input to the following Dense layer. The Softmax function is utilized as activation function in the last layer to evaluate the results as a probability distribution by the provided network. For categorization tasks, the function appears to be really useful. According to Wikipedia, the simplest fundamental formula for the Softmax function is: $\Omega: R^K \rightarrow R^K$, as explained by Wikipedia, is:

$$\Omega(z)_i = \frac{e^{z_i}}{\sum_{j=1}^{K} e^{z_j}} \quad \text{for } i = 1, 2, \ldots, K \text{ and } z = (z_1, z_2, \ldots, z_k) \in R^K \quad\quad\quad (3)$$

Where, the input vector z consists of K real numbers.

For each output layer element, the above method uses a standard exponential function, which is then normalised by dividing by the aggregate of all exponentials. The class of predicted items in the list will be the expected class. In our task, the likelihood of mention's word would fall into the category {SINGLETON(1), NON-SINGLETON (0)}. As shown in Fig. 3, the final output is received when the linearly transformed data is passed through Softmax. The softmax layer determines the probability score of each class. The highest likelihood value is chosen.

### 4. EXPERIMENTAL SETUP AND EVALUATION

We employ this neural network architecture for the Singleton Mention Detection task because of the dataset size limitation imposed by the lack of labeled data. To train our models, we used an NVIDIA 970GTX GPU and a 4.00 GHz Intel i7-4790 processor with 64GB RA with KERAS [48] by utilizing TensorFlow [49] as a backend. The news dataset is randomly divided into two sections in our experiments: training (80%) and test (20%). Validation is performed on 20% of the training set during training.

### 4.1 Experimental Setup

We used Hindi coreference annotated data [50], collected from an online website of IITH. The dataset consists of 3.6K sentences and 78K tokens from news article domains. The singleton mentions in this corpus are not annotated. Therefore, an additional mention detection phase is required to extract singleton mention from the dataset. Our approach employed the Rule-based mention detection algorithm [46], which identified all kinds of anaphors, nominal mention, and verb-nominal sequences as potential mentions. After this step, we scanned all predicted mentions by the algorithm to choose the mentions that are not present in the coreference annotated data set as a singleton mention. After assuming the mentions which are not participating in coreference as a singleton, these mentions are manually annotated as singleton mentions by giving label "Yes" or "No". This corpus is used to create a dataset according to the Singleton Mention Detection task. Table 1 shows the corpus statistics. This corpus is used to create a dataset according to the Singleton Mention Detection task.

*Hyperparameters*

The hyperparameter setting for the presented work is represented in Table 2. This paper employed a similar hyperparameters configuration as utilized by Auliarachman and Purwarianti [36]. The maximum length of mention considered is 10 words. Each word of mention is tokenized before being sent to Word2vec to create a 300-dimensional word vector.

Table 1. Corpus detail

| Domain | Hindi Dataset | Size |
|---|---|---|
| News | # Documents | 275 |
|  | # Sentences | 3.6K |
|  | # Tokens | 78K |

Table 2. Hyperparameters

| Parameter | Value | Parameter | Value |
|---|---|---|---|
| Word Embedding dimension | 300 | FCN layer | 32, 16, 64,8 |
| Dropout | 0.2 | Decay Rate | 0.999 |
| Learning Rate | 0.001 | Epoch | 20 |
| Number of filters (CNN) | 64 | Optimizer | adam |
| Size of filters (CNN) | 2,3,4 | The maximum length of mention | 10 |
| Batch Size | 5 |  |  |

*Word Representation* The embedding layer uses 300-dimensional word embeddings that have been pre-trained on a Wikipedia corpus of 323 M and a vocabulary size of 30393.

*A dropout* is a type of regularization that is used to address the overfitting issue that occurs during model training By placing the Dropout layer before and after the LSTM layers, many links are dropped out at random. For Skip-gram word embedding, the proposed model works well with a dropout rate of 0.2.

*Filters* we are using three distinct filter sizes for each CNN, totaling 64 filters, and also utilized two layers with 32 and 16 dimensions for FCN that are directly linked to the inputs. We suggest three layers with dimensions of 64, 32, and 16 for each FCN after concatenations. Finally, we suggest two 32 and 8 dimensions layers for the final FCN.

We use layers of 2D convolution for the convolution operation, each with sizes 2, 3, and 4 for 64 filters. RELU is utilized as the activation function in all convolution layers.

*Learning* For the learning purposes, the model is optimised with Adam optimizer. The loss is computed using the Sparse Categorical Cross-Entropy loss function. The learning rate is 0.999, while the decay rate is 0.001.

### 4.2 Experimental Results

Using the proposed approach correctly, the system accurately anticipates mentions. The system also uses the following assessment metrics to appropriately report the results:

1. Recall: This refers to the percentage of relevant results that the algorithm properly classifies.

$$Recall = \frac{predicted\ correct\ singleton\ mentions}{Total\ possible\ singleton\ mentions\ in\ gold\ standard} \dots\dots\dots\dots\dots\dots (4)$$

2.  Precision: This refers to the percentage of relevant results that algorithm successfully classifies.

$$Precision = \frac{Predicted\ correct\ singleton\ mentions}{Total\ number\ of\ predicted\ singleton\ mentions} \quad\quad\quad (5)$$

3.  F-Measure: This refers to the weighted average of Precision and Recall.

$$F_\beta = (1 + \beta^2) \frac{PR}{\beta^2 P + R} \quad\quad\quad (6)$$

4.  Accuracy: This refers to the fraction of a number of correct predictions over the total number of predictions. In other words, accuracy indicates the likelihood that our machine learning model will accurately predict a result out of the total number of predictions it has made.

$$Accuracy(p, \hat{p}) = \frac{1}{No.of\ samples} \sum_{i=0}^{No.of\ samples} 1\ (p = \hat{p}) \quad\quad\quad (7)$$

Where $p$ indicate the actual singleton mention of a particular sample and $\hat{p}$ indicate the corresponding predicted singleton mention.

The Singleton Mention Detection module is evaluated with different configurations in this experiment. The experimental result of the presented model with context embedding includes the two previous and two next words of a mention on the dataset with various configurations concerning the Precision, Recall, and F-measure are given in Table 3.

Table 3. Hindi Singleton Mention Detection results with the proposed model with different configurations on the news domain dataset.

| Model | Features | Non-Singleton | | | Singleton | | |
|---|---|---|---|---|---|---|---|
| | | **Precision** | **Recall** | **F-measure** | **Precision** | **Recall** | **F-measure** |
| CNN-FCN | Pretrained word[5] | 74 | 69 | 72 | 62 | 68 | 65 |
| CNN-FCN | Pretrained word + Context[6] | 72 | 74 | 73 | 63 | 61 | 62 |
| CNN-FCN | Pretrained word + Syntactic[7] | 73 | 72 | 73 | 63 | 64 | 65 |
| CNN-FCN | Word + Context + Syntactic | 77 | 69 | 72 | 63 | 72 | 67 |

The Precision, Recall, and F-measure for this configuration of our model are 77% and 69%, 72%, respectively for identifying Non-singleton class. The Precision, Recall, and F-measure for this model configuration are 63% and 72%, 67%, respectively, for identifying the Singleton class. If we do not include the Syntactic feature, then the Precision, Recall, and F-measure for our Singleton Mention Detection model are 73%, 72%, 73%, respectively, for identifying Non-singleton class. The Precision, Recall, and F-measure for this model configuration are 63% and 64%, 65%, respectively for identifying the Singleton class.

---

[5] Pretrained word" refers to a model which covers embedding of pre-trained word embeddings of mention's words.
[6] , "Context " refers to a model which contains embedding of context words of mention (Context feature include only two previous words and two next words of mention)
[7] "Syntactic" refers to model which contains the binary features of mention.

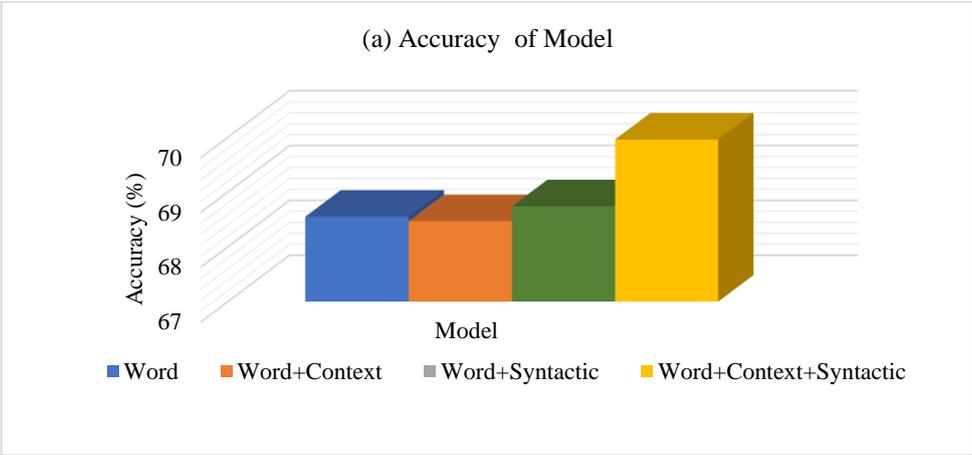

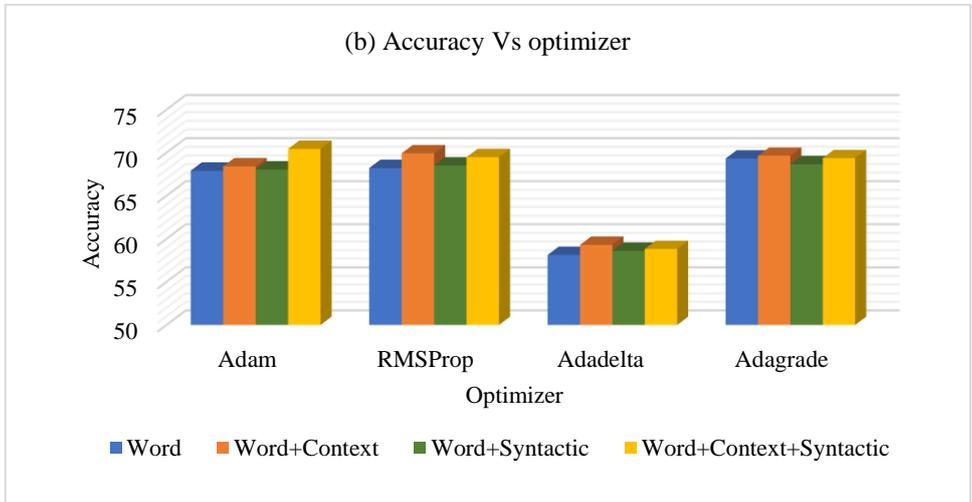

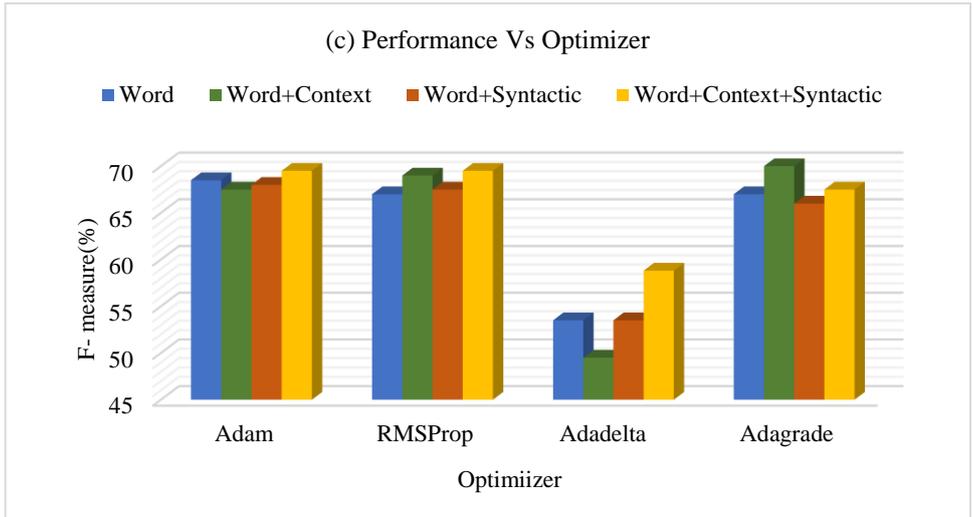

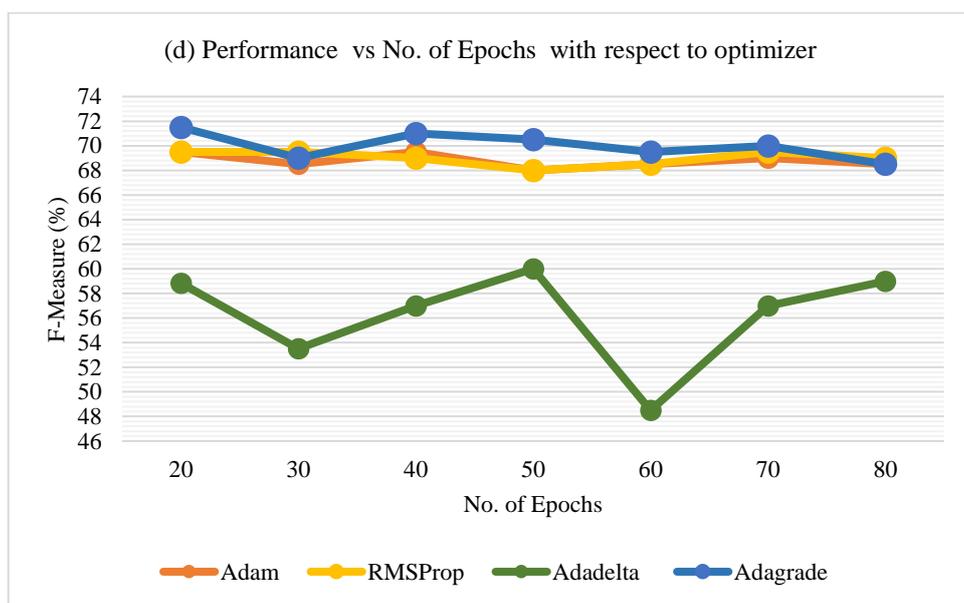

Fig. 5 Performance of Model (Context include two previous words and two next words of mentions) with different Hyper-parameters.

Fig. 5 depicts the change in the proposed model's performance, which includes the context embedding of two previous words and two next words of a mention with different hyper-parameter. From Fig. 5(a), we observe that our word+context+syntactic model achieved an accuracy of 70.45%. Our models had various features to tweak to understand the impact of various aspects on the overall results. Therefore, we investigated the effect of pre-trained word embedding, context embedding, and syntactic features on our CNN-FCN model. It is observed that the use of pre-trained word embedding with context and syntactic features gave us the highest performance over a model with other configurations.

Table 4. Hindi Singleton Mention Detection results with the proposed model with different configurations on the news domain dataset.

| Model | Features | Non-Singleton | | | Singleton | | |
|---|---|---|---|---|---|---|---|
| | | **Precision** | **Recall** | **F-measure** | **Precision** | **Recall** | **F-measure** |
| CNN-FCN | Pretrained word | 73 | 74 | 74 | 64 | 64 | 64 |
| CNN-FCN | Pretrained word + Context[8] | 73 | 71 | 72 | 62 | 65 | 64 |
| CNN-FCN | Pretrained word + Syntactic | 73 | 70 | 72 | 62 | 65 | 63 |
| CNN-FCN | Word + Context + Syntactic | 74 | 72 | 73 | 64 | 66 | 65 |

---

[8] "Context" refers to a model which contains embedding of context words of mention. (Context feature include all previous words and all next words of mention)

. As demonstrated in Fig. 5(b), 5(c), 5(d), the proposed model performs well with the Adam optimizer. We utilize the activation function 'softmax' along with optimizer 'RMSProp,' 'Adam,' 'Adadelta', and 'Adagrad'. The observation made from Fig. 5(d) is that the performance of the model is worst using the 'Adadelta' optimizer. The experimental result of the presented model with context embedding includes all previous words and all next words of a mention on the dataset with various configurations concerning the Precision, Recall, and F-measure are given in Table 4.

The Precision, Recall, and F-measure for this configuration of model (Word + Context + Syntactic) are 74% and 72%, 73% respectively for identifying Non-singleton class. The Precision, Recall, and F-measure for this model configuration are 64% and 66%, 65%, respectively, for identifying the Singleton class. If we do not include the Syntactic feature, then the Precision, Recall, and F-measure for Singleton Mention Detection model are 73%, 71%, 72%, respectively, for identifying Non-singleton class. The Precision, Recall, and F-measure for this configuration of our model are 62% and 65%, 64%, respectively, for identifying the Singleton class.

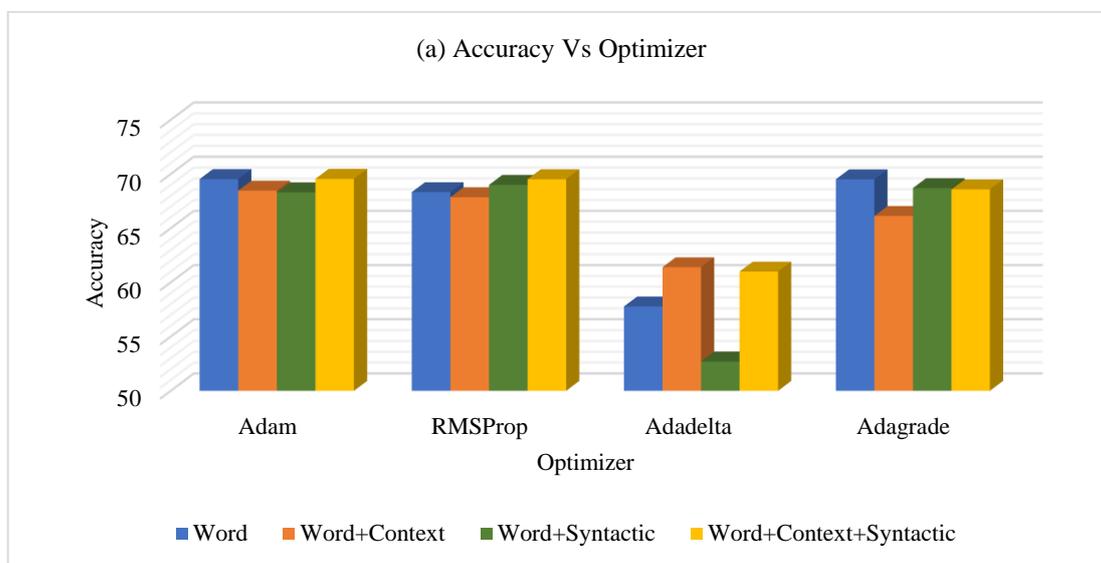

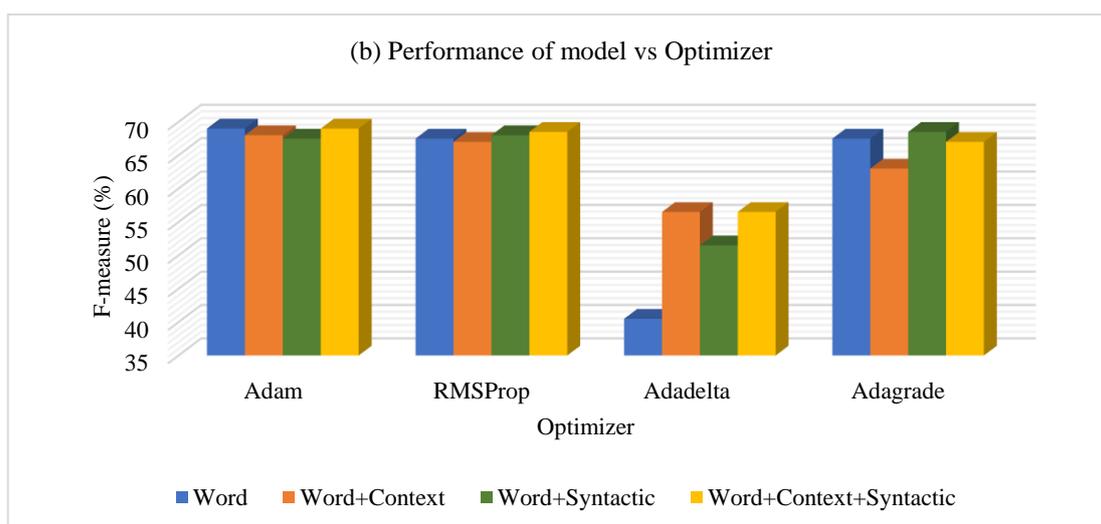

Figure 6 Performance of Model (Context includes all previous words and all next words of mentions) with different Hyper-parameters.

Fig. 6 depicts the performance of the proposed model, which includes the context embedding of all previous words and all next words of a mention with different hyper-parameter. From Fig. 6(a), 6(b), we observe that the word+context+syntactic model achieved an accuracy of 69.55% and performed well with the Adam optimizer. The worst performance of model is by utilizing Adadelta optimizer.

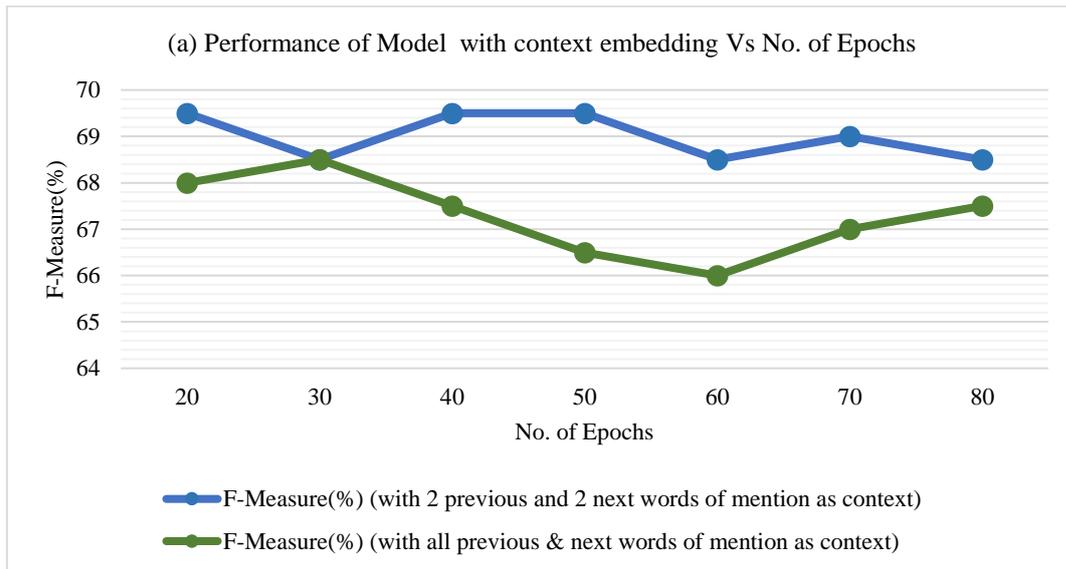

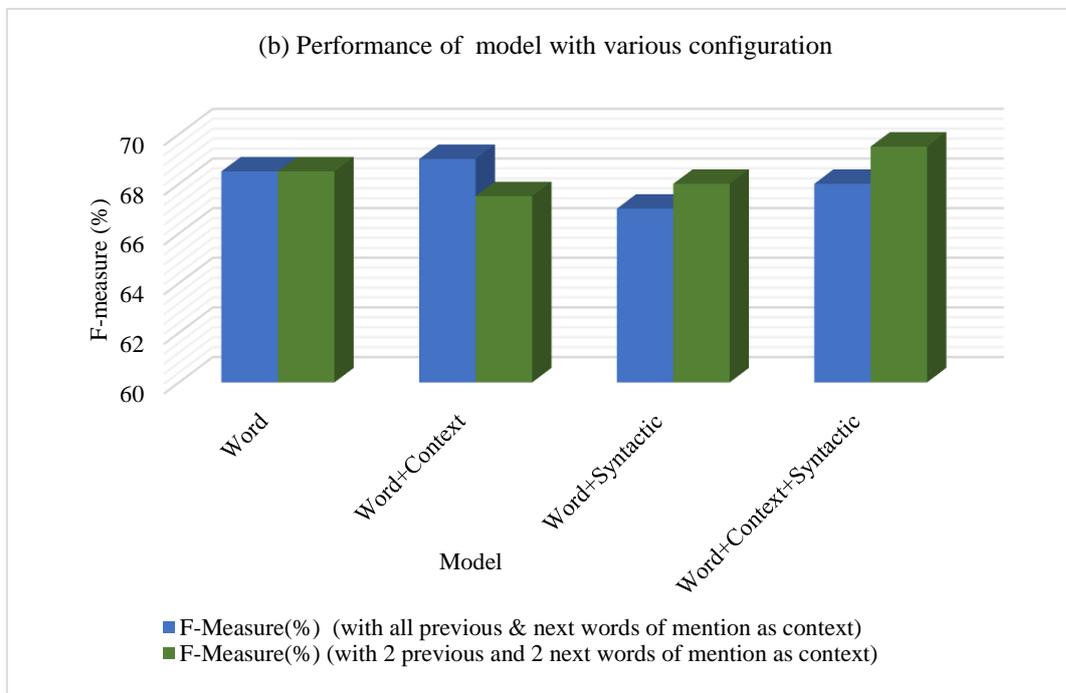

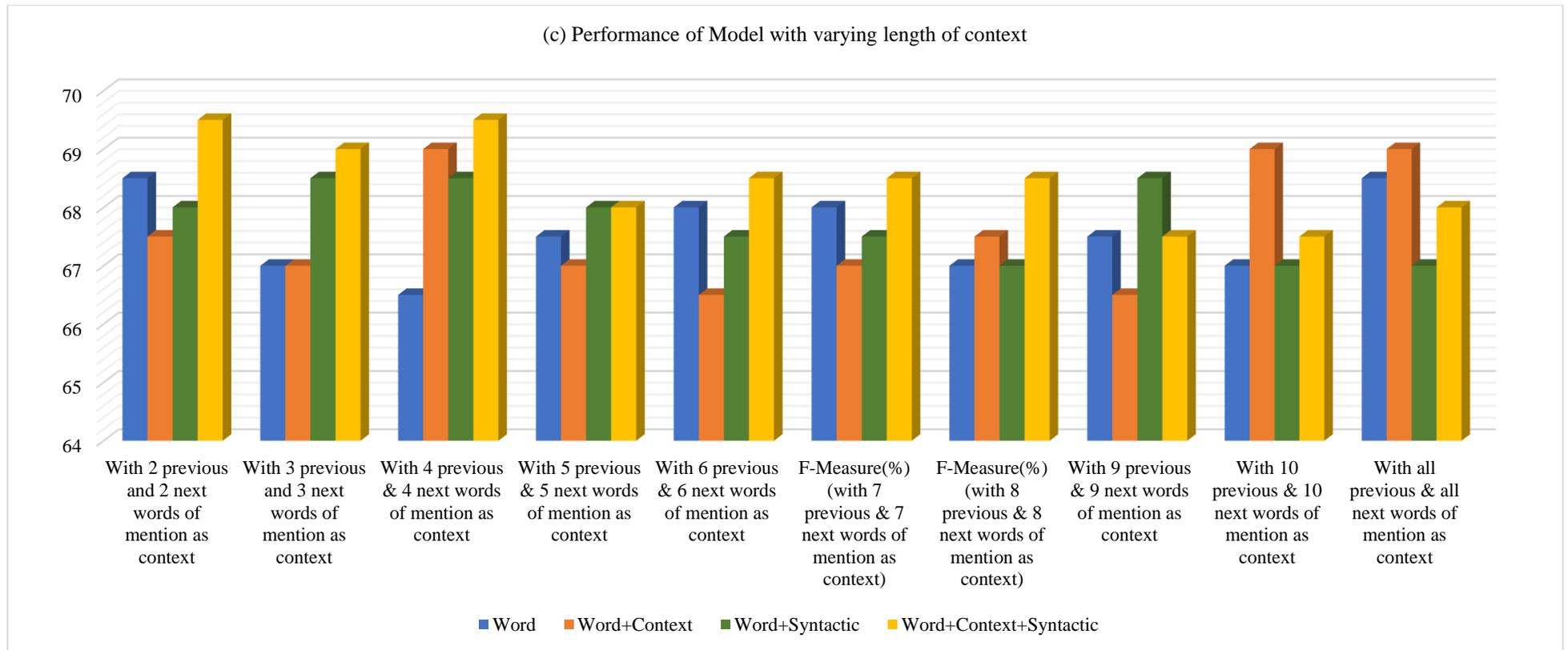

Fig. 7 Performance of Model with different Hyper-parameters

Fig. 7 depicts the comparison of the model's performance with Context embedding includes all previous words and all next words of mention, and our model with Context embedding includes only two previous words and two next words of a mention. As demonstrated in Fig. 7(a), the model's performance varies according to epoch. At epoch 20, the performance of our model (Context embedding includes only two previous words and two next words of a mention) is good. Therefore, no need to increase the number of epoch because it will increase the computation time of model. Fig. 7(b) and 7(c) shows the comparison of proposed model for detecting singletom mentions with varying the length of context of each mention. It is observed from Fig. 7(b) and 7(c) that the proposed model (Context embedding includes only two previous words and two next words of a mention) performs well compared to a model (Context embedding that includes all previous words and all next words of mention). As we increase the size of context window as shown in Fig. 7(c), there is no significant change in the performance of model. We can say that size of context window ( in our case, two is used) is small good to perform better. If we increase the size of context window like "Context embedding includes all previous words and all next words of mention", the computation time and space also increases.

Because the system relies on word embeddings for information, the inclusion of 'unknown words,' i.e., words for which no embedding exists, might jeopardize its performance. We use the CNN for our task because short sequences are used, i-e. Local context. Hindi is a resource-poor language; therefore, a large amount of data set is not available to train neural network models. The dataset we used for our task is small. If the deep learning model learns more, then it will give more accuracy in predicting the output. Therefore, the performance of our Singleton Mention Detection model can be improved by having a large amount of dataset because the deep learning technique requires a large amount of dataset to train and learn. The performance of the deep learning model will improve as more data becomes available.

## 5. CONCLUSION AND FUTURE SCOPE

Singleton Mention Detection is an important step and is utilized by well known NLP application Coreference Resolution which is useful for many NLP application such as Machine Translation, Text Summarization, Reading Comprehension, Entity linking, Question Answering, Sentiment Analysis. The removal of singleton mentions reduces search space for CR tasks and makes the rest of the resolution process easier, faster, and more accurate (fewer mentions to consider). This paper describes a methodology to create a Singleton Mention Detection model to detect singleton mention from the list of mentions extracted from Hindi text. The proposed method is based on deep learning techniques where a Convolutional neural network and a Fully connected network (CNN-FCN) are utilized for the task. We minimize the dependency on hand-crafted features by using this technique.

The proposed model used pre-trained word embedding to acquire the semantic data and a few syntactic features in Hindi text. The result obtained in this paper are based on the Hindi dataset which was created from the scratch. This dataset is relatively small in size and the further studies could be conducted with the larger datasets.